%% file: main.tex
\documentclass[11pt,a4paper]{article}
\usepackage[hyperref]{acl2019}
\usepackage{times}
\usepackage{latexsym}

\usepackage{url}

\usepackage{amsmath}
\usepackage{amssymb}
\usepackage{bbold}
\DeclareMathOperator*{\argmax}{arg\,max}
\usepackage{booktabs}
\usepackage{footmisc}
\usepackage[compact]{titlesec}
\usepackage{graphicx}

\aclfinalcopy 

\title{Ensembling Strategies for Answering Natural Questions}

\author{Anthony Ferritto, Lin Pan, Rishav Chakravarti, Salim Roukos\\
 {\textbf{Radu Florian, J. William Murdock, Avirup Sil\thanks{\ \ Corresponding author.}}}\\ 
IBM Research AI\\
Yorktown Heights, NY\\
aferritto@ibm.com \\ 
\{panl, rchakravarti, roukos, raduf, murdockj, avi\}@us.ibm.com
}

\date{}

\definecolor{darkgreen}{rgb}{0.0, 0.2, 0.13}

\newcommand{\saDevTrain}{$SA\ F1_{Tr}$}
\newcommand{\laDevTrain}{$LA\ F1_{Tr}$}
\newcommand{\saDevTest}{$SA\ F1_{Te}$}
\newcommand{\laDevTest}{$LA\ F1_{Te}$}
\newcommand{\numModelsSearched}{$N_{S}$}
\newcommand{\numModels}{$N_{SA/LA}$}

\newcommand{\fss}[1]{\textbf{#1:}}
\newcommand{\fsss}[1]{\textbf{#1:}}

\begin{document}
\maketitle
\begin{abstract}
  Many of the top question answering systems today utilize ensembling to improve their performance on tasks such as the Stanford Question Answering Dataset (SQuAD) and Natural Questions (NQ) challenges.  Unfortunately most of these systems do not publish their ensembling strategies used in their leaderboard submissions.  In this work, we investigate a number of ensembling techniques and demonstrate a strategy
  which improves our F1 score for short answers on the dev set for NQ by 2.3 F1 points over our single model
  (which outperforms the previous SOTA by 1.9 F1 points).
  





\end{abstract}

\input{introduction.tex}
\input{related-work.tex}
\input{methods.tex}
\input{experiments.tex}
\input{conclusion.tex}



\bibliography{references}
\bibliographystyle{acl_natbib}

\end{document}

%% file: introduction.tex
\section{Introduction}
Machine Reading Comprehension (MRC), a relatively new field in the open domain Question Answering (QA) community, aims to answer a question by reading and understanding a text.  This field has recently received considerable attention, yielding popular leaderboard challenges such as SQuAD \cite{Rajpurkar_2016, rajpurkar2018know} and NQ \cite{Kwiatkowski2019NaturalQA}.  Currently, the top submissions on both the SQuAD and NQ leaderboards utilize ensembling.  These ensembled systems traditionally outperform single models by a couple of F1 points.  Unfortunately, many of the papers for these systems provide little to no information about the ensembling techniques they use. 
In this work, we  expand upon \cite{pan2019frustratingly} to present an ensembling technique that improves upon our SOTA system on the NQ short answer (SA) leaderboard.
Using our ensemble of models, for each example (question, passage pair) we take the top predictions per system, group by span, normalize and aggregate the scores, take the mean score across systems for each span, and then take the highest scoring short and long answer spans as our final prediction.
More completely, our contributions include: \\
\textbf{Ensembling Algorithms:} We present a number of ensembling approaches that can be used to aggregate MRC models. \\
\textbf{Comparison Experiments:} We evaluate these strategies on the NQ dataset to compare their performance.

%% file: related-work.tex
\section{Related Work}



Many of the top recent MRC systems publish few details on their ensembling strategies.  The system from \cite{Devlin2018BERTPO,albert-synth-data} uses a six model ensemble which adds approximately 1.4 F1 points on the SQuAD 2.0 test set.
RoBERTa \cite{roberta} uses a five to seven model ensemble for GLUE \cite{wang2018glue}, however they do not detail the performance gains from doing so.
Unfortunately neither of these systems provide in depth information on their ensembling approaches.
ALBERT \cite{lan2019albert} gains 1.3 points on the SQuAD 2.0 test set with an ensemble of 6 to 17 models by averaging scores for spans with multiple probabilities (and also averaging ``unanswerability'' scores).
Other systems such as  XLNet \cite{xlnet} mention that they use ensembling but do not provide further details.  MSRA's R-Net \cite{group2017r-net} uses an ensemble of 18 identically trained model, selecting the answer with the highest sum of confidence scores at inference for an improvement of around 2.5 F1 points on the SQuAD 1.0 test set.  BiDAF \cite{DBLP:journals/corr/SeoKFH16-bidaf} uses the same technique with 12 models for an improvement of 4 F1 points on the SQuAD 1.0 test set.

We also consider work in the field of information retreival (IR) as inspiration for our ensembling methods, as a way to aggregate multiple scores for the same span.  Similar to the popular CombSUM and CombMNZ \cite{inproceedings-combsum-combmnz, Wu:2012:DFI:2222499-combsum-combmnz} methods, considering the spans as the ``documents'', we use span-score weighted aggregation in our noisy-or aggregator.   Futher, we additionally incorporate the use of rank-based scoring from Borda \cite{YOUNG197443-borda} and RRF \cite{Cormack:2009:RRF:1571941.1572114-rrf} for our exponential sum approach (in addition to utilizing score for this approach).  We finally consider a reciprocal rank sum aggregation strategy based on the ideas in RRF \cite{Cormack:2009:RRF:1571941.1572114-rrf}.


%% file: methods.tex
\section{Methods}
\label{sec:ensemble}

We investigate a number of strategies for ensembling models.  In order to formally compare approaches we partition the 
NQ dev set 
into ``dev-train'' and ``dev-test''
by taking the first three dev files for the ``train'' set and using the last two for the ``test'' set  (the original dev set for NQ is partitioned into 5 files for distribution).  This yields ``train'' and ``test'' sets of 4,653 and 3,177 examples respectively 
.  For each straregy considered we search for the best k-model ensemble over the ``train'' set and then evaluate on the ``test'' set.  For these experiments we use $k=4$ as this is the number of models that we can decode in the 24 hours on a Nvidia\textsuperscript{\textregistered} Tesla\textsuperscript{\textregistered} P100 GPU, which is the limit for the
NQ leaderboard\footnote{
This was the submission hardware available at the time of our submission in early September 2019.
The available hardware has since been changed to 2 Nvidia\textsuperscript{\textregistered} Tesla\textsuperscript{\textregistered} V100 GPUs}.  We begin by outlining our core strategy that underlies the approaches we have investigated.  Using this strategy we investigate a baseline approach of ensembling multiple versions of the same model trained with different seeds in addition to a number of search, normalization, and aggregation strategies and the impact they have on F1 performance.

\fss{Core Strategy}
For each example processed by the $k$ systems being ensembled, our system assigns a score to each long and short span\footnote{Note that our system currently only predicts single short spans rather than sets, so we currently score each short span independently.} according to the normalization and aggregation strategies (see below).  We use the top-20 candidate long and short answers (LA and SA respectively) for each system and example as we have empirically found this to perform better than considering fewer candidates (e.g. 5 or 10)\footnote{We have found emperically that using top-20 is ideal as an accuracy/runtime tradeoff given hardware resources.}. 
To combine systems we take the arithmetic mean\footnote{We have experimented with other approaches such as median, geometric mean, and harmonic; however these are omitted here as they resulted in much lower scores than arithmetic mean.} of the scores for each long and short span predicted by at least one system.  For spans which are only predicted by some systems a score of zero is assigned (for the systems which do not predict the span) to penalize spans which are only predicted by some systems. The predicted long span is then the span with the greatest arithmetic mean.  Similarly for short answers the predicted span is the one with the greatest arithmetic mean, with the exception that it may not be contained with in a null long span. 

\fss{Seed Ensembles}
We first examine the baseline approach of ensembling $k$ versions of the same model trained with the same hyperparameters, only varying the seed between models.  We select the model based on \cite{pan2019frustratingly} with the highest sum of short and long answer F1 scores on dev.  These models are then ensembled using the core strategy.


\fss{Search Strategies}
We consider two main strategies when searching for ensembles: exhaustive and greedy\footnote{We also considered a ``simple greedy'' approach where the k best models on dev were selected, however this underperformed other pproaches by 1 - 2 F1 points}.  In exhaustive search we consider all possible ensembles, whereas in greedy search we build the ensemble one model at a time by looking for which model we can add to an $i$ model ensemble to make the best $i+1$ model ensemble.



\fsss{Exhaustive Search (ES)}
In the exhaustive search approach where we consider each of the ${n \choose k}$ ensembles of k candidates from our group of n models.
We then use our core strategy for each ensemble to obtain short and long answer F1 scores for each ensemble.
After searching all possible ensembles we return two ensembles: (i) the ensemble with the highest long answer F1 score and (ii) the ensemble with the highest short answer F1 score.



\fsss{Greedy Search (GS)}
We then consider the greedy approach.  In order to more precisely control the tradeoff between optimizing for short and long answers we add an additional parameter $0 \leq k_{S} \leq k$ to specify the number of models which should be used to optimize short answer performance (i.e. $k = k_{S}$ indicates to exclusively optimize for short answers).  This induces another parameter $k_{L} = k - k_{S}$, the number of models which are used to optimize for long answer performance.  We refer to these ensembles that will be created for short and long answers as $S$ and $L$ respectively.  

We construct S by greedily building $1, 2, ..., k_{S}$ model ensembles optimizing for short answer F1 using our core strategy.
In case adding some of the models decreased our short answer performance, we take the first $i \leq k_{S}$ models of $S$ which give the highest short answer F1.  Similarly we construct $L$ by greedily building a $k_{L}$ model ensemble optimizing for long answer F1.

Since we are already decoding all of the models in $L$, we check to see if adding any subset of these models to $S$ improves short answer performance.  More formally we create $S^{\prime} = S \cup \argmax_{x \in \mathcal{P}(L)} F1_{S}(S \cup x)$ where $F1_{S}(X)$ is the short answer F1 for the ensemble created with the models in $X$.  We do the same for optimizing long answer performance, creating $L^{\prime} = L \cup \argmax_{x \in \mathcal{P}(S)} F1_{L}(L \cup x)$. 

Finally, we join the predictions for short and long answers together by taking the short answer and long answer predictions from $S^{\prime}$ and $L^{\prime}$ respectively.  If for an example a null long answer is predicted, we also predict a null short answer regardless of what $S^{\prime}$ predicted as there are no short answers for examples which do not have a long answer in NQ \cite{Kwiatkowski2019NaturalQA}.

\fss{Normalization Strategies}
We investigate two primary methods for normalizing the scores predicted for a span: not normalizing and logistic regression\footnote{We also investigated normalizing by dividing the scores for a span by the sum of all scores for the span, however we omit these results for brevity as they did not produce interesting results.}.


\fsss{None}
As a baseline we run experiments where the scores for a span are used as-is.

\fsss{Logistic Regression}
We also experiment with normalization using logistic regression where the scores from the top prediction for the ``dev-train'' examples is used to predict whether the example is correctly answered\footnote{In our experiments using the top example performed equally well to using the top 20 predictions per example to train on.  We also experimented with using other features which did not improve performance.}.  To ensure an appropriate regularization strength is used, we use the scikit-learn \cite{scikit-learn} implementation of logistic regression with stratified 5-fold cross-validation to select the L2 regularization strength.


\fss{Aggregation Strategies}
We consider a number of aggregation strategies to produce a single span score for each span predicted by a system for an example.  These include the baseline approache of max as well as the exponentially decaying sum, reciprocal rank sum, and noisy-or methods influenced by IR.  Note that all of these approaches operate on a vector $P$ of scores on which one of the above normalization strategies has been applied.

\fsss{Max}
For the max method we take the greatest score produced by a system for a span, or more formally for a vector $P$, $score = \max_{i=1}^{|P|} P_{i}$.





\fsss{Exponential Sum (ExS)}
For exponential sum, based on the ideas of \cite{YOUNG197443-borda, Cormack:2009:RRF:1571941.1572114-rrf}, we reverse sort $P$ and take $score = \sum_{i=1}^{|P|} P_{i} * \beta^{i - 1}$ for some constant $\beta$ (we use $\beta = 0.5)$.

\fsss{Reciprocal Rank Sum (RRS)}
For reciprocal rank sum, based on the ideas of \cite{Cormack:2009:RRF:1571941.1572114-rrf}, we reverse sort $P$ and take $score = \sum_{i=1}^{|P|} P_{i} * \frac{1}{i}$.

\fsss{Noisy-Or (NO)}
For noisy-or, based on the ideas of \cite{inproceedings-combsum-combmnz, Wu:2012:DFI:2222499-combsum-combmnz}, we take $score = 1 - \prod_{i=1}^{|P|} (1 - P_{i})$.

\begin{figure}[h]
\caption{Ensembling system overview.  Group by (GB) collects score predictions by span.  Algorithm abberviations detailed in section \ref{sec:ensemble}.}
\centering
\includegraphics[width=0.35\textwidth]{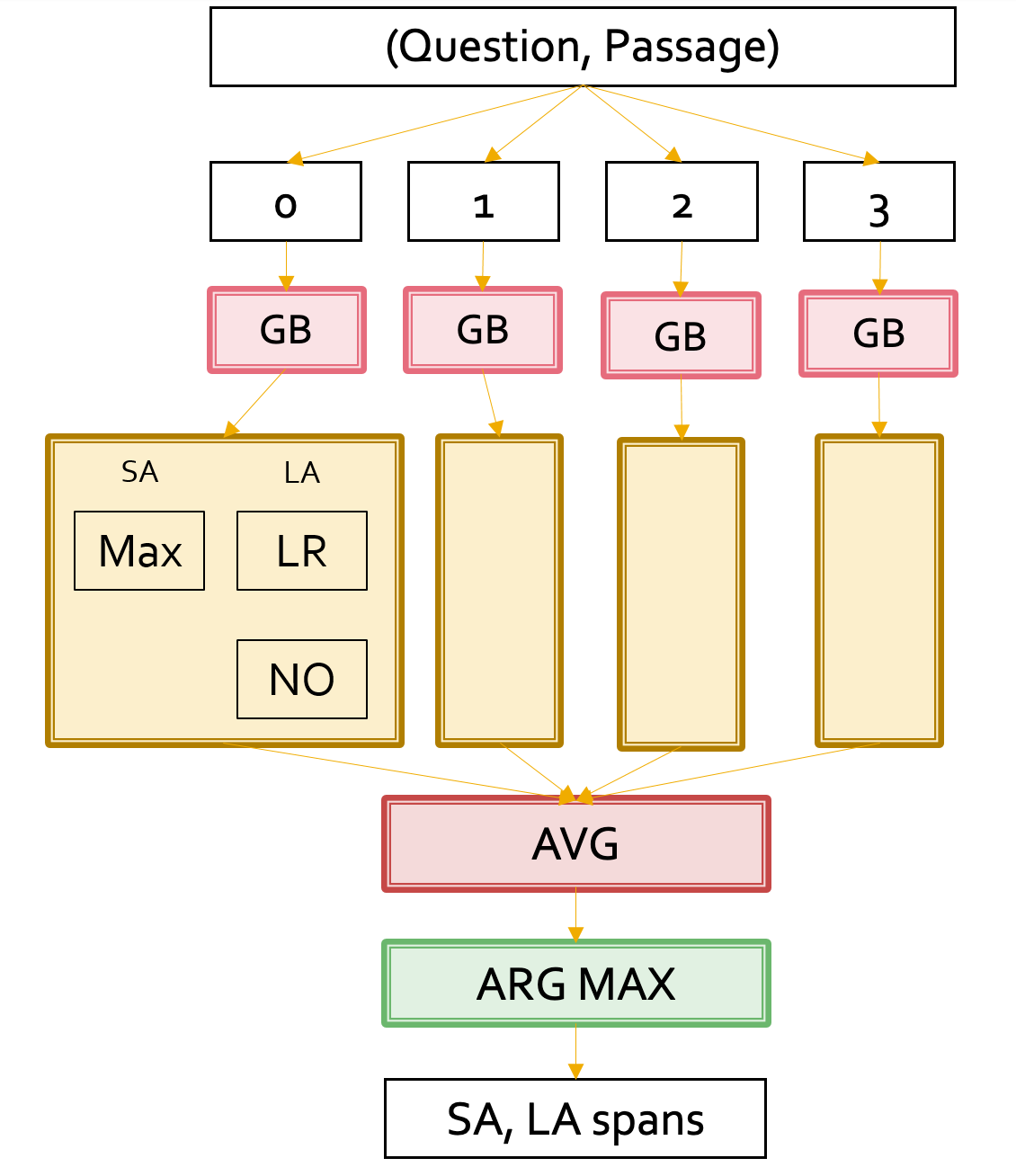}
\label{fig:overview}
\end{figure}

In figure \ref{fig:overview} we show an overview of our ensembling system, using non-normalized max for short answers and logistic regression normalized noisy-or for long answers.

%% file: experiments.tex
\section{Experiments}
We examine two types of ensembling experiments: (i) ensembling the same model trained with different seeds and (ii) ensembling different models.  Ensembling the same model trained on different seeds attempts to smooth the variance to produce a stronger result.  On the other hand ensembling different models attempts to find models that may not be the strongest individually but harmonize well to produce strong results.
Throughout this section we will use \saDevTrain, \laDevTrain, \saDevTest, \laDevTest \ to denote the short and long answer performance on dev train and test.  Similarly we will use \numModelsSearched \ to indicate the number of models searched for an experiment and \numModels \ to indicate the number of models an experiment uses to optimize short and long answer F1 performance.




\fss{Seed experiments}
In table \ref{tab:exp:seed} we find that there is a benefit to ensembling multiple versions of the same model trained with different seeds\footnote{Note that there is some data snooping ocuring here as models are selected based on full dev performance (which is a superset of ``dev-test'').}.


\begin{table}[]
\small
\centering
\begin{tabular}{lll}
\toprule 
\# Models  & \saDevTest & \laDevTest   \\
\toprule 
1  & 0.5614 & 0.6710  \\
\hline
4  & \textbf{0.5873} & \textbf{0.6961} \\
\bottomrule 
\end{tabular}
\caption{Ensembling the same model trained with different seeds.}
\label{tab:exp:seed}
\end{table}

\begin{table*}[]
\small
\centering
\begin{tabular}{llllllll}
\toprule
 Search & \numModelsSearched & \numModels  & \saDevTrain & \laDevTrain & \saDevTest & \laDevTest   \\
 \toprule
 ES & 20 & 0/4 & 0.5922 & \textbf{0.6989} & 0.5964 & 0.7049   \\
 \hline
 ES & 20 & 4/0 & \textbf{0.5925} & 0.6902 & 0.5964 & 0.6998   \\
 \hline
 GS & 41 & 0/4 & 0.5857 & 0.6972 & \textbf{0.5971} & 0.7084  \\
 \hline
 GS & 41 & 1/3 & 0.5767 & 0.6929 & 0.5863 & 0.\textbf{7097}    \\
 \hline
 GS & 41 & 2/2 & 0.5837 & 0.6932 & 0.5864 & 0.7089    \\
 \hline
 GS & 41 & 3/1 & 0.5883 & 0.6853 & 0.5857 & 0.7060    \\
 \hline
 GS & 41 & 4/0 & 0.5897 & 0.6922 & 0.5907 & 0.6981   \\
 \bottomrule
\end{tabular}
\caption{Comparison of Search Strategies.  All experiments run without normalization using the max aggregator.}
\label{tab:exp:ens:search}
\end{table*}

\fss{Main experiments}
We investigate the different search strategies in table \ref{tab:exp:ens:search}.
For the exhaustive approach we see that it obtains the best long and short F1 scores on ``dev-train'' for the respective models as expected, however these do not translate to the best scores on ``dev-test''.  We find that the greedy approach performs best overall, with the greedy ensemble optimized exclusively for long answer performance performing the best on short answer F1 and is 0.0013 F1 lower than the best long answer F1 on ``dev-test''.  Also note that the numbers seen here, particularly when optimizing greedily for long answer performance are higher than those observed for ensembling the same model with multiple seeds in table \ref{tab:exp:seed}.

We hypothesize that the reasons for the superior generalization of the greedy approach over exhaustive is that exhaustive search is ``overfitting'' and that greedy can search more candidates.  Whereas with the greedy approach we can search all 41 candidates\footnote{Candidate models represent models trained with different model types (e.g. BERT for QA and AoA) and hyperparameters (e.g. learning rate and negative example subsampling rate)}, with exhaustive search we only consider our top 20 (by sum of short and long answer F1) for runtime considerations.  This gives the greedy strategy a more diverse set of candidates.  Similarly we hypothesize the reason optimizing for long answer F1 generalizes better for short and long answers is due to the strict definition of correctness for Natural Questions which requires exact span matching \cite{Kwiatkowski2019NaturalQA}.  For the remainder of this paper we will focus on greedy search with all four models optimized for long answer to keep the number of experiments presented to a manageable level \footnote{Note that further experimentation with different short and long answer optimization trade-offs confirm the conclusion here that optimizing for long answer performance generalizes the best.}.


\begin{table}[]
\small
\centering
\begin{tabular}{lllll}
\toprule
\numModelsSearched  & \saDevTrain & \laDevTrain & \saDevTest & \laDevTest   \\
\toprule
41  & 0.5857 & \textbf{0.6972} & \textbf{0.5971} & \textbf{0.7084}  \\
\hline
20  & \textbf{0.5860} & 0.6951 & 0.5896 & 0.704    \\
\bottomrule
\end{tabular}
\caption{Impact of candidate pool size.  All experiments run with a greedy search strategy optimized exclusively for long answer F1 without normalization using the max aggregator.}
\label{tab:exp:ens:candidates}

\vspace{1.5em}

\small
\centering
\begin{tabular}{lllll}
\toprule
Agg.  & \saDevTrain & \laDevTrain & \saDevTest & \laDevTest   \\
\toprule
Max  & \textbf{0.5857} & 0.6972 & \textbf{0.5971} & 0.7084  \\
\hline
ExS  & 0.5619 & 0.6944 & 0.5826 & 0.7040  \\
\hline
RRS  & 0.5553 & 0.6954 & 0.5728 & 0.7066  \\
\hline
NO  & 0.5545 & \textbf{0.7037} & 0.573 & \textbf{0.715}  \\
\bottomrule
\end{tabular}
\caption{Comparison of IR inspired aggregation strategies (Agg.).  All experiments run with a greedy search strategy optimized exclusively for long answer F1 with logistic regression normalization (except max which is not normalized).}
\label{tab:exp:ens:norm-agg}





\end{table}

In table \ref{tab:exp:ens:candidates} we investigate the impact of the candidate model search space size.  We use the greedy search strategy to optimize for long answer performance without normalization using the max aggregator.  For the experiment with 20 candidates we use the same 20 candidates as for the exhaustive search strategy experiment (greatest sum of short and long answer F1s).  We find that the added diversity of the models avaible in the entire pool produces a stronger ensemble which generalizes better both in short and long answer F1.  Having verified the benefit of using the entire candidate pool we continue to do so in the remainder of our experiments.

We investigate the impact of the IR inspired normalization strategies in table \ref{tab:exp:ens:norm-agg}.  The max experiment is as-before run without normalization to greedily optimize for long answer F1.  The other experiments here are normalization with logistic regression, as our experiments showed that not normalizing decreased performance.  We find that using max aggregation results in the best short answer F1 whereas using normalized noisy or aggregation results in the best long answer F1.  Based on these results, we run a final experiment using unnormalized max for short answers and logistic regression normalized noisy or works for long answers.  We find that this approach produces the strongest performance for both short and long answers with 0.5934 \saDevTest \hspace{0.25pt} and 0.7150 \laDevTest.  These numbers translate to a full dev performance of 0.5933 short answer F1 and 0.7107 long answer F1, which represents an improvement of 2.3 short answer F1 and 4.0 long answer F1 over our best single model.

%% file: conclusion.tex
\section{Conclusion}
We outline several ensembling approaches for question answering models and compare their performance on the NQ challenge.  We find that ensembling unique models outperforms ensembling the same model trained with different seeds.  We also show that using unnormalized max aggregation for short answers and logistic regression normalized noisy or aggregation for long answers yields an F1 improvement of 2 to 4 points over single model performance on the NQ challenge.